\documentclass{article}

\usepackage{times}
\usepackage{graphicx} 
\usepackage{subfigure} 

\usepackage{natbib}

\usepackage{algorithm}
\usepackage{algorithmic}

\usepackage{mathtools, amsmath, amsfonts, amssymb, amsthm}

\theoremstyle{remark}

\numberwithin{equation}{section}

\theoremstyle{remark}

\usepackage{hyperref}

\usepackage[accepted]{icml2016}

\icmltitlerunning{Balancing Method for High Dimensional Causal Inference}

\begin{document} 

\twocolumn[
\icmltitle{Anomaly Detection in Bitcoin Network Using \\ 
          Unsupervised Learning Methods}

\icmlauthor{Thai T. Pham}{thaipham@stanford.edu}
\icmlauthor{Steven Lee}{slee2010@stanford.edu}

\icmlkeywords{Anomaly detection, unsupervised learning.}

\vskip 0.3in
]

\begin{abstract} 
The problem of anomaly detection has been studied for a long time. In short, anomalies are abnormal or unlikely things. In financial networks, thieves and illegal activities are often anomalous in nature. Members of a network want to detect anomalies as soon as possible to prevent them from harming the network's community and integrity. Many Machine Learning techniques have been proposed to deal with this problem; some results appear to be quite promising but there is no obvious superior method. In this paper, we consider anomaly detection particular to the Bitcoin transaction network. Our goal is to detect which users and transactions are the most suspicious; in this case, anomalous behavior is a proxy for suspicious behavior. To this end, we use three unsupervised learning methods including $k$-means clustering, Mahalanobis distance, and Unsupervised Support Vector Machine (SVM) on two graphs generated by the Bitcoin transaction network: one graph has users as nodes, and the other has transactions as nodes.
\end{abstract} 

\section{Introduction}
Network structures have appeared for a long time, and along with them are those who behave abnormally within the system. We refer to these people or their illegal activities as anomalies. With respect to financial transactional networks, anomalies can include those who execute fraudulent transactions. In these networks, a common goal is to detect those anomalies to prevent future illegal actions. 

Bitcoin is a special type of transaction system; more information about it can be found in \url{https://bitcoin.org/en/}. We seek to detect anomalies or suspicious activities in this anonymous network, where nodes (i.e. users, transactions) are unlabeled and there is no confirmation as to whether or not a given node is actually conducting illicit activities. 

In this project, we focus particularly on the problem of detecting anomalies in the Bitcoin transaction network, which is related to the study of fraud detection in all types of financial transaction systems in which a large literature exists \cite{FPA2016, PL2016}. Since this problem can be generalized to those in other network settings, which may or may not involve financial transactions, we are examining the more general problem of anomaly detection in networks. 

In this paper, we use three unsupervised learning methods including $k$-means clustering, Mahalanobis distance based method, and Unsupervised Support Vector Machines (SVM) on two graphs generated by the Bitcoin transaction network: one graph has users as nodes, and the other has transactions as nodes. 

The rest of the paper is organized as follows. Section $2$ describes everything related to methods, including data collection and parsing, feature extraction, and mathematical descriptions of the machine learning techniques. Section $3$ discusses evaluation methods for our proposed algorithms. Section $4$ presents the results we obtained by running these techniques on the network-type data set we generated. Section $5$ evaluates our methods and results. Section $6$ describes future work. Section $7$ concludes our study. 

\section{Methods}
In this section, we describe data collection and parsing. Then, we describe feature extraction, and finally we provide mathematical explanations for the unsupervised learning techniques we use.

\subsection{Data Collection and Parsing}
We use the Bitcoin transaction data set obtained from the University of Illinois Urbana- Champaign. All Bitcoin transactions are documented in a public ledger and are in the currency unit called the Bitcoin (BTC). The data set contains all Bitcoin transactions beginning from the network’s creation until April 7th, 2013. For each transaction, there can be multiple sender and receiver addresses. Furthermore, multiple addresses can belong to a single user. Finally, users are also anonymous in that there are no names or personal information associated with a given user. 

The data set is quite large: there are $6,336,769$ users with $37,450,461$ transactions. We parse the data in two ways. The first way, which we will call the \emph{user graph}, is quite intuitive: users (where each user owns a list of addresses) are nodes and transactions between users are edges. The second way, which we will call the \emph{transaction graph}, models transactions as nodes and Bitcoin flow between transactions as edges.

In our analysis, we will use both graph types to investigate the Bitcoin network. The user graph will help us detect suspicious users, while the transaction graph will help us detect suspicious transactions. Using these two graph representations, we can not only find out both abnormal users and abnormal activities, but also check if our methods are consistent in the sense that suspicious transactions should belong to suspicious users.

\subsection{Feature Extraction}
In order to use $k$-means as a baseline, calculate the Mahalanobis distance, use an Unsupervised SVM, for each node in the graph we extract a set of features. In each of the graph representations of the data (mentioned above), we extract the following $12$ features. Keep in mind that even with the same names, the features mean different things for the two representations. 

\begin{itemize}
  \item In-degree, out-degree, unique in-degree, unique out-degree, clustering coefficient
  \item Average in-transaction, average out-transaction
  \item Average time interval between in-transactions, average time interval between out-transactions, 
  \item Balance, creation date, active duration
\end{itemize}

\subsection{Unsupervised Learning Techniques}

\subsubsection{$k$-means Clustering}

The purpose of the $k$-means clustering method is to partition $m$ points (i.e. $m$ nodes in the graph) into $k$ groups of similar characteristics. Technically speaking, $k$-means clustering itself is not a method for anomaly detection; however, it can be useful. Because we expect outliers to stay far away from the centroids found by $k$-means, $k$-means can be used to assess our true methods.

For this method to work, we first represent each node as a multi-dimensional vector in the Euclidean space; each dimension of a node is a feature that we choose from the list described in part $2.2$. For runtime purposes, we select only a subset of features. For the user graph, we use six features: in-degree, out-degree, mean incoming transaction value, mean outgoing transaction value, mean time interval, and clustering coefficient. For transaction graph, we use three features: in-degree, out-degree, total amount of each transaction.

This method produces a set of $m$ points $(x_1, ..., x_m)$
in which $x_i \in \mathbb{R}^n$ (where $n = 6$ or $3$ depending on graphs) for each $i = 1, ..., m$. We seek to partition these $m$ points into $k$ clusters $S = (S_1, ..., S_k)$ to solve 
\begin{equation*} \min_S \sum_{i = 1}^k \sum_{x \in S_i} ||x - \mu_i||^2, \end{equation*}
where $\mu_i$ is the mean of the points in $S_i$ for each $i = 1, ..., k$. We use the $k$-means clustering algorithm as a heuristic method to solve this problem. The algorithm in details can be found in \cite{L1982}. Our final note in this part is that we use the normalized log of feature values to account for different value scales of different features.  

\subsubsection{Mahalanobis Distance Based Method}
This method is based on the Multivariate Gaussian Distribution assumption. 

Specifically, we assume the training set $(x_1, ..., x_m)$ where $x_i \in \mathbb{R}^n$ (again, $n = 6$ for user graph and $n = 3$ for transaction graph) drawn from multivariate normal distribution
\begin{equation*} p(x; \mu, \Sigma) = \frac{1}{(2\pi)^{\frac{n}{2}} |\Sigma|^{\frac{1}{2}}} \exp\left(-\frac{1}{2} (x - \mu)^T \Sigma^{-1} (x - \mu)\right). \end{equation*}

The normal parameter fitting method (MLE) implies that $\mu$ and $\sigma$ can be estimated by
\begin{equation*} \widehat{\mu} = \frac{1}{m} \sum_{i = 1}^m x_i \;\textit{ and }\; \widehat{\Sigma} = \frac{1}{m} \sum_{i = 1}^m (x_i - \mu)(x_i - \mu)^T. \end{equation*}

Then, we will flag $x$ as an anomaly if $p(x, \widehat{\mu}, \widehat{\Sigma}) < \epsilon$ for some chosen threshold $\epsilon$. Note that our data are unlabelled, so we will train and detect on the same data set.

\subsubsection{Unsupervised SVM}
The usual SVM method does not work here because our data are unlabeled, so we will use a modified version of it. We also take advantage of the Kernel trick so that we can use an infinite dimensional feature space. We also assume the training set $(x_1, ..., x_m)$ where $x_i \in \mathbb{R}^n$ for all $i$. 

We start with the primal optimization problem:
\begin{equation*} \min_{w, \rho} \left(\frac{1}{2} w^T w\right) \; \textit{s.t.} \; w^Tx_i \geq \rho \; \forall i = 1, ..., m. \end{equation*}
A new test example $x$ will be labeled $sign(w^T x - \rho)$. The point $x$ with label $(-1)$ is flagged as an anomaly.  

Note that the data are not linearly separable, so we need to use a soft-margin SVM. However, we will not use the constant $C$ as in the usual soft-margin SVM due to its difficulty in interpretation. Instead, we follow \cite{SPSSW2001} to use $\nu$-SVM method as follows.

\begin{equation*} \min_{w, \xi, \rho} \left(\frac{1}{2} w^T w + \frac{1}{\nu m} \sum_{i = 1}^m \xi_i - \rho\right), \textit{ s.t.} \end{equation*}
\begin{equation*} w^T x_i \geq \rho - \xi_i \; \forall i = 1, ..., m \textit{ and } \xi_i \geq 0 \; \forall i = 1, ..., m. \end{equation*}
Here, $\nu \in (0, 1]$ can be interpreted as an upper bound for the fraction of outliers (i.e. anomalies). In the Results section, we will determine which $\nu$ is optimal.  

Following \cite{SPSSW2001} again, we can write the corresponding dual problem using the Kernel trick:
\begin{equation*} \min_{\alpha} \left(\frac{1}{2} \sum_{i = 1}^m \sum_{j = 1}^m \alpha_i \alpha_j K(x_i, x_j) \right) \textit{ subject to } \end{equation*}
\begin{equation*} 0 \leq \alpha_i \leq \frac{1}{\nu m} \; \forall i = 1, ..., m \textit{ and } \sum_{i = 1}^m \alpha_i = 1. \end{equation*}
Here, we choose $K(x, z) = \exp(-\gamma(x - z)^2)$ where the hyper-parameter $\gamma$ will be fine-tuned later. 

To solve for this optimization problem efficiently, we can use SMO method \cite{P1998}.

Finally for $0 < \alpha_j < \frac{1}{\nu m}$, we can recover $\rho$:
\begin{equation*} \rho = \sum_{i = 1}^m \alpha_i K(x_i, x_j). \end{equation*}
Then a new point $x$ is flagged anomaly if $\sum_{i = 1}^m \alpha_i K(x_i, x) < \rho$. 
Again, since our data are all unlabelled, we will will use the same data set for training and detecting. 

\section{Evaluation Methods}
With unlabeled data, evaluating our methods is a difficult challenge. Due to the nature of the network-type data set we have, we propose three evaluation methods.
\begin{itemize}

    \item[\textbullet] Using $k$-means as a baseline, we can calculate the relative distances between the detected outliers and the centroids. If these values are small, then we conclude that our methods are not good enough. We call this ``Visualization Evaluation.'' 
    
    \item[\textbullet] Since we represent our data in two ways with nodes and edges somehow exchanging, we can test for our methods' consistency by checking if detected suspicious users own detected suspicious transactions. We call this ``Dual Evaluation.''   
    
    Specifically, with the user graph we can get the top $N$ user outliers and with the transaction graph we can get the top $M$ transaction outliers. In this paper, we choose $N = M = 100$. We then determine $X_N$ - the set of transactions corresponding to the top $N$ node outliers and $Y_M$ - the set of users corresponding to the top $M$ transaction outliers defined above. We define
    
    \begin{equation*} A_1 = \frac{|X_N \cap \text{ top $X_N$ transaction outliers}|}{|X_N|} \end{equation*}
and
    \begin{equation*} A_2 = \frac{|Y_M \cap \text{ top $Y_M$ user outliers}|}{|Y_M|}. \end{equation*}

Finally, we define the Dual Evaluation Metric $m_{DE}$ by
    \begin{equation*} m_{DE} = \frac{A_1 + A_2}{2}. \end{equation*}
Note that $m_{DE} \in [0, 1]$, and the bigger it is the more accurate our method is.
    
    \item[\textbullet] Finally, there are roughly $30$ revealed thieves in the Bitcoin network. We can check if they belong to our detected suspicious user and transaction sets. For this reason, these users and their illegal transactions will be included in the test sets of our methods.
    
\end{itemize}

\section{Results}
The $\nu$-SVM method takes a long time to run, and since we are not able to use techniques like GPU-parallelized computation at the moment, we will limit our data set to $100,000$ data points for all methods. 

\subsection{k-Means Clustering}
Using the $k$-means clustering metric in \cite{XWC2006}, we find that setting $k = 7$ minimizes cross-cluster entropy for the user graph and $k = 8$ for the transaction graph. (Figure $1$.) For the sake of the dual evaluation method, we choose $k = 7$ for both graph types.  

\begin{figure}[h]
\begin{center}
\includegraphics[scale=0.5]{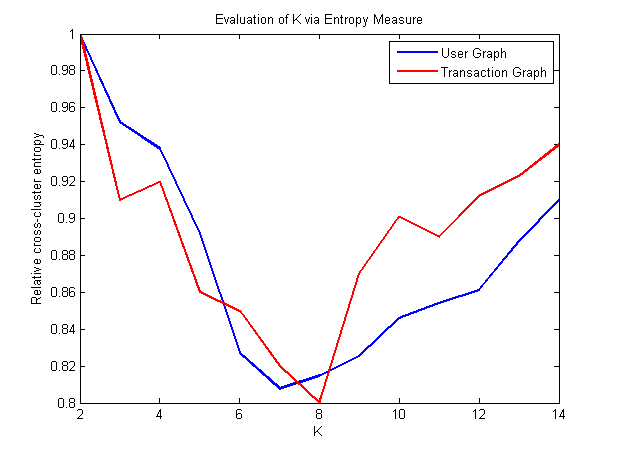}
\caption{Cluster Entropy vs. k}
\end{center}
\end{figure}

\subsection{Mahalanobis Distance Based Method}
Now, we run the Mahalanobis distance based method for two types of graphs. (Figures $2-3$.)

\begin{figure}[h]
\begin{center}
\includegraphics[scale=0.4]{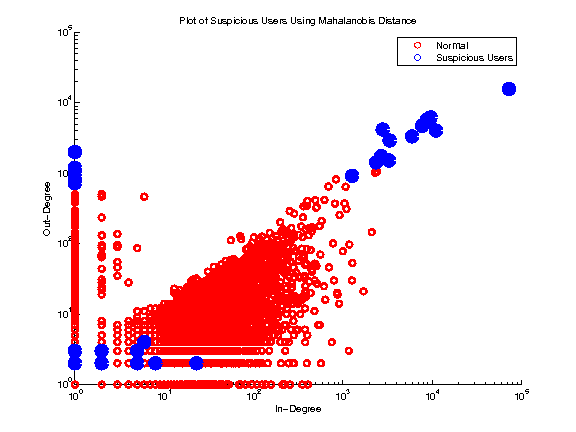}
\caption{Anomaly Detection using Mahalanobis Distance: User Graph}
\end{center}
\end{figure}

\begin{figure}[h]
\begin{center}
\includegraphics[scale=0.4]{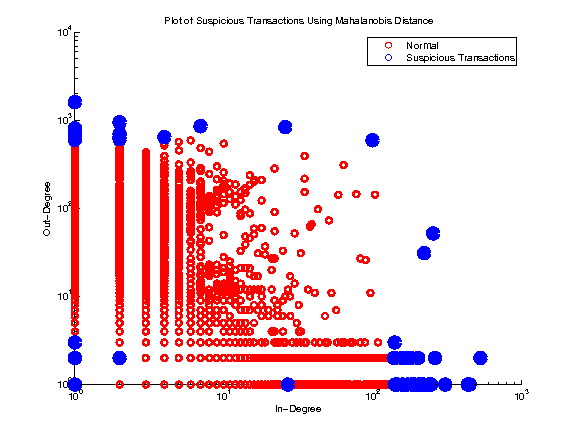}
\caption{Anomaly Detection using Mahalanobis Distance: Transaction Graph}
\end{center}
\end{figure}

The detected anomalies seem to appear at the border of the plot, which indicates that abnormal activities are usually extreme. 
 
\subsection{Unsupervised SVM}
We now run the Unsupervised (one - class) $\nu$-SVM method for two graph types. We start with identifying the optimal hyper-parameter $\nu$. Based on the evaluation section, we will choose the $\nu$ which gives the largest values of $A_1$ and $A_2$. It turns out that the optimal $\nu$ is around $0.005$. (Figure $4$.)

\begin{figure}[h]
\begin{center}
\includegraphics[scale=0.3]{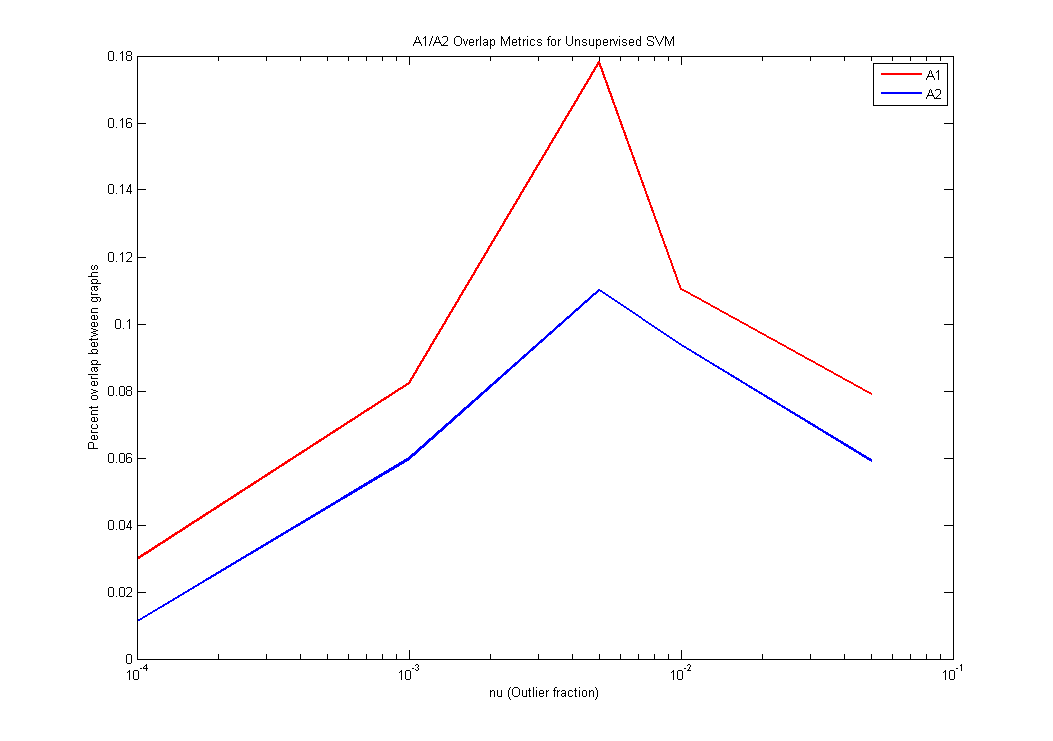}
\caption{Plot of $A_1$ and $A_2$ over $\nu$ in $\nu$-SVM method.}
\end{center}
\end{figure}

Now with $\nu = 0.005$, the $\nu$-SVM method gives us the detected anomalies for both types of graph representations in Figures $5$ and $6$. 

\begin{figure}[h]
\begin{center}
\includegraphics[scale=0.4]{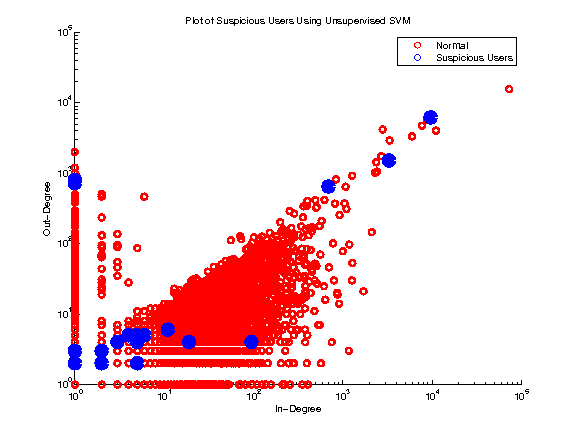}
\caption{Anomaly Detection using Unsupervised SVM: User Graph}
\end{center}
\end{figure}

\begin{figure}[h]
\begin{center}
\includegraphics[scale=0.4]{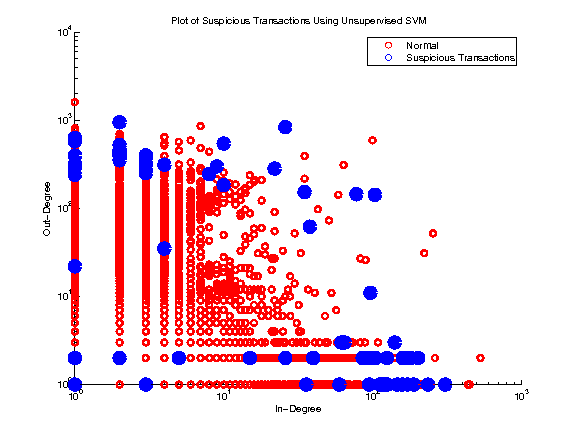}
\caption{Anomaly Detection using Unsupervised SVM: Transaction Graph}
\end{center}
\end{figure}

The suspicious users and transactions suggested by the Unsupervised SVM method appear to be quite similar to those suggested by the Mahalanobis distance based method. They both usually appear on the outer borders of the graphs.

\section{Evaluation Results}
\begin{itemize}

    \item[\textbullet] Using $k$-means clustering method, we get $k$ clusters with corresponding $k$ centroids. For each graph type (user and transaction) and for each method (Mahalanobis and Unsupervised SVM), we calculate the average of the ratios of detected anomaly distances to corresponding centroids over max distances from those centroids to their assigned points for the top $100$ outliers. 
    
    For the Mahalanobis method, we get $0.7619$ for the user graph and $0.8277$ for the transaction graph. For the Unsupervised SVM method, we get $0.7192$ for the user graph and $0.8584$ for the transaction graph. These values are as large as we expect them to be since by observations in the Result section, detected anomalies appear to be extreme points. 
    
    \item[\textbullet] For the Mahalanobis Distance based method, we calculate $A_1$ and $A_2$ to be $0.02495$ and $0.026316$ respectively, which gives $m_{DE} = 0.025633$. Though this value is quite small, it comes out not as a surprise given the simplicity of our method and our small set of data. 
    
    For the Unsupervised SVM method, we get $A_1 = 0.1782$ and $A_2 = 0.1101$ which gives $m_{DE} = 0.14415$. This value is much higher than the one produced by the Mahalanobis distance based method, even given our small set of data.

    \item[\textbullet] For the Mahalanobis Distance based method, we detect one known theft that occurred in June 2011.  The anomalous transaction obtained a total of over 4000 BTC from 620 various addresses and funneled them to a single address.
    
For the Unsupervised SVM based method, we detect one known loss that occurred in October $2011$. The anomalous transaction was one of 23 transactions that caused a user to lose over $2,600$ BTC due to corruption in a hashing function. 
    
\end{itemize}

\section{Future Studies}
We propose to parallelize computation in order to enable faster outlier detection using a GPU.  This involves significant work, such as introducing thread-safety in the learning methods, but will allow us to analyze the full dataset of $\sim 38$ million transactions.

\section{Conclusions}
In this paper, we have investigate the Bitcoin network. We first represent the data with two focuses: users and transactions. We then use three main social network techniques to detect anomalies, which are potential anomalous users and transactions. While the agreement metrics are not high, we are able to detect two known cases of theft and one known case of loss, out of the 30 known cases we have. 

{\normalsize{
\bibliography{anomaly_unsupervised}
\bibliographystyle{icml2016}
}}

\end{document}